\pdfoutput=1

\documentclass[11pt]{article}

\usepackage[preprint]{acl}

\usepackage{times}
\usepackage{latexsym}

\usepackage{CJKutf8}
\usepackage{amsmath}
\usepackage{makecell}
\usepackage{subfig}
\usepackage{multirow}
\usepackage{anyfontsize}
 \usepackage{booktabs} 

\usepackage[T1]{fontenc}

\usepackage[utf8]{inputenc}

\usepackage{microtype}

\usepackage{inconsolata}

\usepackage{graphicx}

\usepackage{amssymb}
\usepackage{amsmath}

\usepackage{graphicx}

\title{Dialogue Language Model with Large-Scale Persona Data Engineering}

\author{Mengze Hong$^{1,2}$ \hspace{0.6ex} Chen Jason Zhang$^{1}$ \hspace{0.6ex} Chaotao Chen$^{2}$ \hspace{0.6ex} Rongzhong Lian$^{2}$ \hspace{0.6ex} Di Jiang$^{2}$\thanks{Di Jiang is the corresponding author: \href{mailto:dijiang@webank.com}{dijiang@webank.com}}\\
$^{1}$Hong Kong Polytechnic University \quad $^{2}$AI Group, WeBank Co., Ltd
}

\begin{document}
\maketitle
\begin{abstract}
Maintaining persona consistency is paramount in the application of open-domain dialogue systems, as exemplified by models like ChatGPT. Despite significant advancements, the limited scale and diversity of current persona dialogue datasets remain challenges to achieving robust persona-consistent dialogue models. In this study, drawing inspiration from the success of large-scale pre-training, we introduce PPDS, an open-domain persona dialogue system that employs extensive generative pre-training on a persona dialogue dataset to enhance persona consistency. Specifically, we present a persona extraction model designed to autonomously and precisely generate vast persona dialogue datasets.  Additionally, we unveil a pioneering persona augmentation technique to address the invalid persona bias inherent in the constructed dataset. Both quantitative and human evaluations consistently highlight the superior response quality and persona consistency of our proposed model, underscoring its effectiveness.
\end{abstract}

\section{Introduction}
\label{sec1}

The open-domain dialogue systems have gained significant interest due to their wide industrial applications, such as customer service support \cite{song2021smartsales, similar_question_generation}, virtual assistance \cite{MandamadiotisKE21}, and social chatbots \cite{zhou-etal-2024-characterglm, NG2025104068}. Inspired by the recent success of large-scale pre-training in natural language processing, many neural dialogue models resort to pre-training on large-scale dialogue datasets \cite{xu-zhao-2021-dialogue} and demonstrate substantial progress in open-domain dialogue. Notable examples include DialoGPT \cite{ZhangSGCBGGLD20}, SPACE \cite{10.1145/3477495.3532069}, and Blender \cite{shuster2022blenderbot3deployedconversational} for English dialogue, as well as CDial-GPT \cite{WangKZHJZH20}, PLATO-2 \cite{bao-etal-2021-plato}, and EVA \cite{gu2023eva2} for Chinese.

\begin{figure}
    \centering
    \includegraphics[width=0.95\columnwidth]{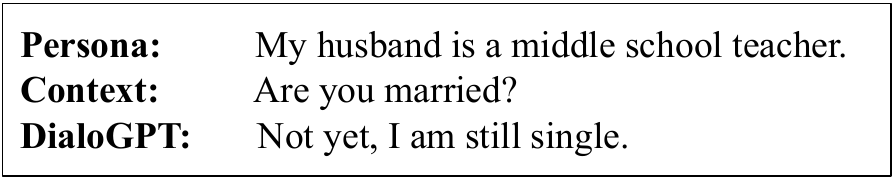}
    \caption{Example of a persona-inconsistent response generated by DialoGPT.}
    \label{fig:badcse}
    \vspace{-0.5em}
\end{figure}

While these methods can generate fluent and coherent responses, maintaining context consistency - particularly persona consistency - remains a common challenge that can lead to negative experiences in real-world human-bot interactions \cite{WelleckWSC19, an2024mcllavamulticonceptpersonalizedvisionlanguage}. As illustrated in Figure \ref{fig:badcse}, the well-trained DialoGPT model struggles with maintaining persona consistency, i.e., revealing contradicted marriage status and generating out-of-character responses. Similar issues appear more frequently with the emerging LLM-based chatbots such as ChatGPT and Claude, where although the chatbot can comprehend user intention correctly \cite{hong-dial-in-llm}, the response behavior often deviates from the instruction prompt and results in user dissatisfaction \cite{tseng-etal-2024-two, dam2024completesurveyllmbasedai, song2024communication}.

To enhance user experience and preserve persona consistency, many research efforts propose introducing explicit personas into the dialogue model \cite{QianHZXZ18,WuMY21}. 
For example, the PERSONA-CHAT dataset \cite{KielaWZDUS18} represents persona through personality sentences. While such crowd-sourced datasets capture a variety of persona features, their small scale, which is limited by the high cost of annotation, prevents them from fully unlocking the potential of large-scale neural dialogue models. On the other hand, the Personality Assignment Dataset \cite{QianHZXZ18} leverages persona attributes from users' social media profiles to automatically create a significantly larger persona dataset. However, the persona diversity is still limited by the attribute set of user profiles. Some studies impose persona consistency using Natural Language Inference (NLI), but their effectiveness is still limited by domain mismatches with general NLI datasets or the scale of dialogue-specific NLI datasets \cite{WelleckWSC19}.


In this paper, we present an efficient solution for constructing large-scale and diverse persona dialogue data, based on which we further pre-train an open-domain persona dialogue model called \textbf{PPDS} (\textit{\textbf{P}re-trained \textbf{P}ersona \textbf{D}ialogue \textbf{S}ystem}) to achieve persona consistency. To construct the dataset, we propose a persona extraction model based on the existing dialogue NLI dataset \cite{WelleckWSC19}, using a summarization approach to automatically and accurately extract personas from large-scale dialogue datasets, such as Reddit comments \cite{BaumgartnerZKSB20}. Strict filtering rules have been implemented to ensure the quality of the persona dataset. Next, we train a large-scale Transformer-based model on the persona dialogue dataset, aiming to enhance its persona consistency through extensive pre-training. Finally, we conduct extensive quantitative and human evaluations to verify the superiority of our model. The contributions of this paper are summarized as follows:

\begin{enumerate}
    \item We propose a persona extraction model to automatically construct large-scale persona dialogue datasets from existing dialogue corpora. Experiments on Reddit comments present a significantly larger and more diverse dataset than current public datasets built from user profiles or human annotations.
    \item We develop a novel open-domain dialogue system pre-trained on the constructed large-scale persona dialogue dataset to enhance persona consistency. A new persona augmentation technique is introduced to address the persona bias issue in the dataset.
    \item Extensive experiments involving both quantitative and human evaluations demonstrate the effectiveness of our model compared to various baselines. We analyze the roles of pre-training, persona augmentation, and fine-tuning as key components, providing insights and best practices for industrial deployment.
\end{enumerate}

\section{Related Work}
\label{sec:related}

\subsection{Large-Scale Pre-Training}
Large-scale pre-training has been a popular paradigm in natural language processing. 
With large Transformer \cite{VaswaniSPUJGKP17} model pre-training in massive plain texts and fine-tuning in downstream tasks, it has demonstrated substantial improvement and generality \cite{DevlinCLT19,LiuOGDJCLLZA19}. 
Recent attempts for larger models and data sizes further reveal the increasing potential of large-scale pre-training. Particularly, the GPT-3 \cite{BrownMRSKDNSSAA20} model with 175 billion parameters demonstrates strong zero-shot and few-shot learning capacities without task-specific fine-tuning on downstream tasks.

Motivated by the efficacy of large-scale pre-trained language models such as GPT-3 \cite{BrownMRSKDNSSAA20}, UniLM \cite{DongYWWLWGZH19} and T5 \cite{RaffelSRLNMZLL20}, many recent efforts in dialogue try to build open-domain dialogue systems through large-scale pre-training on human-like dialogue. Equipped with large amounts of dialogue data collected from social media such as Reddit, Twitter, Weibo, etc, these models can generate human-like responses and enhance the engagingness of human-AI conversations. 
Although these methods have achieved substantial enhancements in open-domain dialogue, they still suffer from the consistency problem, especially persona consistency \cite{RollerDGJWLXOSB21,NieWBKW21}.

\subsection{Persona Dialogue Model}
To solve the problem of persona consistency, recent works focus on a data-driven approach where a persona dialogue dataset is introduced to capture the persona-related features. The persona include user identity \cite{LiGBSGD16}, user profiles \cite{QianHZXZ18} and persona facts \cite{KielaWZDUS18,MazareHRB18}. 
To leverage the persona information, many well-designed neural models are proposed, such as modeling mutual-persona \cite{LiuCCLCZZ20} and multi-stage persona-based dialogue generation \cite{SongWZLL20}. 
Besides, there also exist many works \cite{abs-1901-08149,GolovanovKNTTW19,abs-1901-09672,RollerDGJWLXOSB21,LinMBF21} demonstrating that fine-tuning pre-trained models on persona dataset can obtain substantial improvement on persona consistency. 
However, due to the limitation of scale and diversity of the public persona dialogue dataset, these methods are still far from achieving satisfactory persona consistency.

In addition to capturing persona consistency implicitly, some works turn to explicitly imposing persona consistency by natural language inference (NLI). 
With an NLI model to judge whether a response contradicts the personas, the dialogue models are able to improve their persona consistency by reranking \cite{WelleckWSC19}, unlikelihood training \cite{LiRKWBCW20,SongWZZL21} or reinforcement learning \cite{SongZH020}.

\section{Large-scale Persona Dialogue Dataset}
\label{sec:dataset}
\subsection{Persona Extraction Model}
To construct the large-scale persona dialogue dataset, we first build a persona extraction model.
Following \cite{WelleckWSC19}, we represent a persona as a triple, i.e. $\boldsymbol{p}=\{\boldsymbol{e}_{1}, \boldsymbol{r}, \boldsymbol{e}_{2}\}$, where $\boldsymbol{e}_{1}$, $\boldsymbol{e}_{2}$ and $\boldsymbol{r}$ denote the subject, object and persona attribute respectively, e.g. $(i, like, swimming)$. 
In particular, we propose to model the persona extraction problem as a summarization task, where the persona triple can be "summarized" from the utterance.
Formally, given an utterance $\boldsymbol{R}$, the persona extraction model outputs the corresponding persona $\boldsymbol{p}$ in a manner of generative summarization by considering the persona triple as a text $\boldsymbol{e}_{1}\ [SEP]\ \boldsymbol{r}\ [SEP]\ \boldsymbol{e}_{2}$, where the delimiter $[SEP]$ is used to distinguish each element in the persona triple. 
For utterances that are irrelevant to persona, we use a special token $[None]$ as their summarization, following the setting in \cite{WelleckWSC19}.

\begin{figure}
\centering
\includegraphics[width=0.95\columnwidth]{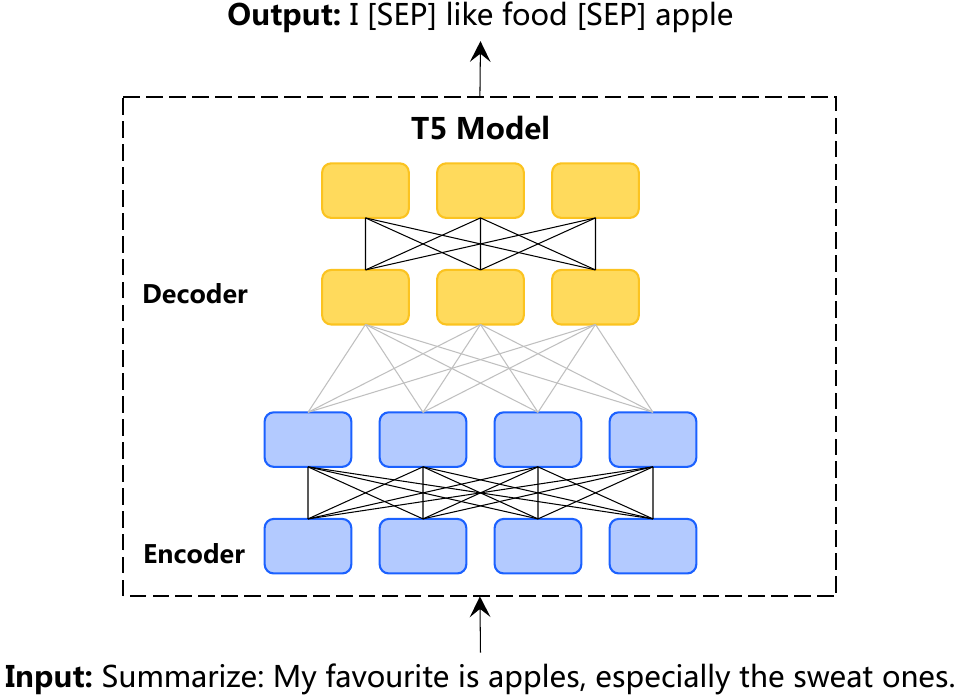}
\caption{Overview of Persona Extraction Model.}
\label{fig:persona_extraction_model}
\end{figure}

The overview of the persona extraction model is illustrated in Figure \ref{fig:persona_extraction_model}.
Specifically, we leverage the Text-to-Text Transfer Transformer (T5) \cite{RaffelSRLNMZLL20} pre-training model as the backbone of our persona extraction model. 
T5 combines many language problems into a text-to-text format for multi-task learning, achieving superior performance in summarization tasks. It is also regarded as a cost-efficient model in the current landscape, making it suitable for industrial deployment.

We employ the Dialogue NLI (DNLI) dataset \cite{WelleckWSC19} as the training corpus for persona extraction.
The DNLI dataset is built upon the PERSONA-CHAT \cite{KielaWZDUS18} dataset by manually annotating persona triple for each utterance.
The details of the persona attributes can be referred to in the original paper \cite{WelleckWSC19}. 
Compared with the dataset presented in \cite{QianHZXZ18}, whose personas are limited by the attribute set of user profiles, the proposed method can capture more diversified personas. 
We fine-tune the T5-large model on the DNLI dataset. Ultimately, the persona extraction model achieves a ROUGE-L score of 80.0\% on the DNLI test set, demonstrating its effectiveness in summarizing personas from utterances.


\begin{table}
    \centering
    \resizebox{\columnwidth}{!}{
    \begin{tabular}{ccccc}
        \toprule 
        \#Session & \#Utterance & \#Persona & \#Token & \#Token/Utterance \tabularnewline
        
        \midrule
        
        189M & 470M & 36M & 12B & 25.5 \tabularnewline
        
        \bottomrule
    \end{tabular}}
    \caption{Statistics of the constructed large-scale persona dialogue dataset.}
    \label{tab:Data_Statistics}
\end{table}

\subsection{Data Construction}
To build the large-scale persona dialogue dataset, we employ the well-trained persona extraction model to automatically extract the persona from the utterances in Reddit comments \cite{BaumgartnerZKSB20}, which consists of 5,601,331,385 comments.
After extracting the persona of each utterance, the following summarized personas are removed to ensure persona quality:


\begin{itemize}
    \item Personas that do not follow the format "$\boldsymbol{e}_{1}\ [SEP]\ \boldsymbol{r}\ [SEP]\ \boldsymbol{e}_{2}$";
    \item Personas with attributes outside of the predefined set of persona attributes;
    \item Personas whose subject exceeds 5 tokens;
    \item Personas with semantic cosine similarity to the original utterance below 0.1, as measured by the sentence-transformer library \cite{ReimersG19}.
\end{itemize}

Finally, we merge the personas from the same character in a dialogue session as a persona profile.
Table \ref{tab:Data_Statistics} shows the statistics of the constructed large-scale persona dialogue dataset. 
To the best of our knowledge, this dataset is the largest of its kind, featuring a diverse range of personas beyond the scale of any existing datasets. It is also worth noting that the scale can be further expanded by leveraging a large dataset of utterances.

\section{Large-Scale Pre-Training}
\subsection{Model}
Based on the constructed large-scale persona dataset, we pre-train a Transformer-based dialogue model PPDS.
Formally, let $\boldsymbol{C}=\{\boldsymbol{c}_{1}, \boldsymbol{c}_{2}, ..., \boldsymbol{c}_{N}\}$ 
denotes the dialogue context which consists of $N$ utterances, $\boldsymbol{R}$ denotes the target response, and 
$\boldsymbol{P}=\{\boldsymbol{p}_{1}, \boldsymbol{p}_{2}, ..., \boldsymbol{p}_{M}\}$ 
denotes the personas which consists of $M$ triples of persona. 
The target of the proposed model $M$ is to generate a persona consistent response $\hat{\boldsymbol{R}}$ based on both persona $\boldsymbol{P}$ and dialogue context $\boldsymbol{C}$, i.e., $\hat{\boldsymbol{R}} = M(\boldsymbol{C}, \boldsymbol{P})$.

\begin{figure}[htbp]
    \centering
    \includegraphics[width=0.9\columnwidth]{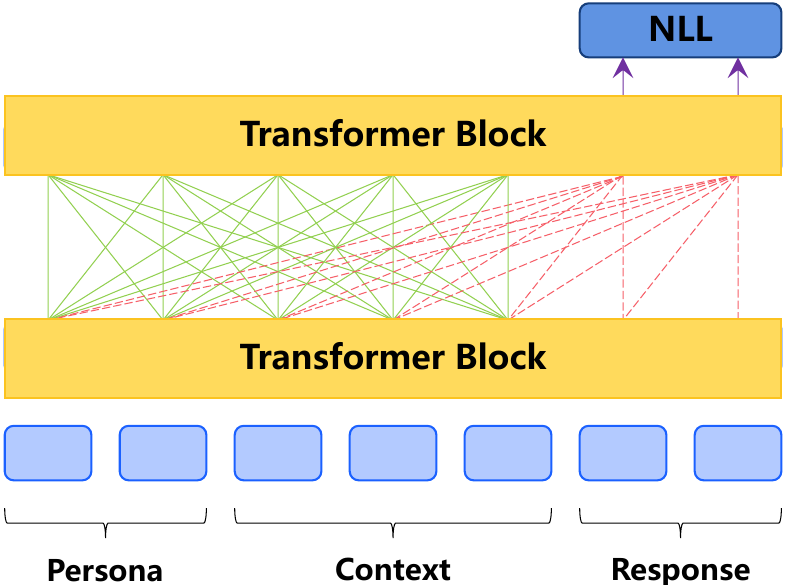}
    \caption{Network architecture of PPDS.}
    \label{fig:architecture}
\end{figure}

The network architecture of PPDS is illustrated in Figure \ref{fig:architecture}. Similar to the existing pre-training dialogue model, it employs Transformer blocks as the backbone.
In order for efficient training on large-scale datasets, PPDS adopts the unified Transformer (also known as UniLM \cite{DongYWWLWGZH19}) instead of the typical encoder-decoder architecture for dialogue generation.
By concatenating the persona, dialogue context, and response as a single input, the UniLM architecture can significantly reduce unnecessary computation of padding. The flexible mechanism of the self-attention mask can also simultaneously model the two tasks of dialogue context understanding and response generation with sharing parameters. 
Therefore, the UniLM architecture is more parameter-efficient than the encoder-decoder network \cite{bao-etal-2021-plato}. Additionally, UniLM has demonstrated promising performance across various downstream tasks \cite{10.1145/3503161.3548112, 10.5555/3524938.3524998}, highlighting its superiority and suitability.

\begin{figure}
    \centering
    \includegraphics[width=0.95\columnwidth]{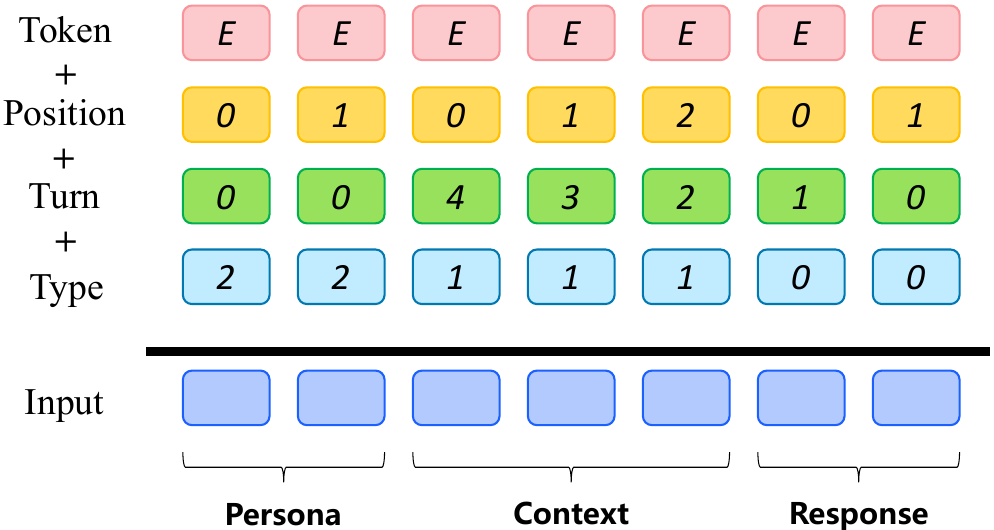}
    \caption{Input representation of PPDS.}
    \label{fig:model_input}
\end{figure}

As shown in Figure \ref{fig:model_input}, the model input is the concatenation of persona, dialogue context, and response. 
Its representation is calculated as the sum of the corresponding token, position, turn, and type embeddings. 
The $token$ is a BPE token in English input or a character in Chinese input.
The $position$ is the index of the token in an utterance. 
The $turn$ is the turn distance of the utterance in dialogue context toward the target response.
We assume that the closer utterances would be more relevant to the target response. 
Specifically, the turn index of target response and persona is $0$.
The $type$ is used to distinguish the characters in the dialogue, where $0$ refers to the responder (i.e., bot), $1$ refers to the respondent (i.e., human), and $2$ refers to the persona profiles.

In detail, the utterances in dialogue context are separated by a special token $[SEP]$.
The persona is represented by the concatenation of persona triple (i.e., "$\{head\}\ \{relation\}\ \{tail\}$"), and each persona is also separated by a special token $[SEP]$.
Once we get the input representations, the UniLM Transformer will perform multi-head attention \cite{VaswaniSPUJGKP17} on the input to transform the embeddings into a sequence of hidden representations $\boldsymbol{H}$. 
Finally, we leverage \textit{Softmax} to transform the hidden representations into the predictive probability of the target response \cite{jiang2023probabilistic, 9422195}.
The details of the Transformer structure can refer to \cite{DongYWWLWGZH19}.

\subsection{Training Objectives}
In PPDS, the pre-training objective is to minimize the widely adopted negative log-likelihood (NLL) loss as follows:

\begin{align}
    \mathcal{L}_{NLL} 
    &= -E_{(\boldsymbol{P},\boldsymbol{C},\boldsymbol{R})}[log\ p_{\theta}(\boldsymbol{R}|\boldsymbol{P},\boldsymbol{C})] \\
    &= -E_{(\boldsymbol{P},\boldsymbol{C},\boldsymbol{R})}[log\ p_{\theta}(r_{t}|\boldsymbol{P},\boldsymbol{C},r_{<t})]
\end{align}

\noindent where $\theta$ refers to the trainable parameters, $T$ is the length of the target response and $r_{<t}$ denotes previously generated words.

\subsection{Persona Augmentation}
Since the large-scale persona dataset is built by extracting personas from the responses, there exists a persona bias in the raw persona dataset. Characters with extracted personas are often linked to persona-related responses, potentially misleading the model to generate such responses whenever a persona profile is present, regardless of its relevance to the dialogue context.

To mitigate this bias, we propose augmenting the dialogue with unrelated personas, compelling the model to identify the relevant persona based on the dialogue context. Specifically, we collect all extracted personas and sample some to supplement each dialogue's persona profiles. If the sampled persona is of the same type as the existing one, we remove it to avoid contradictions. Finally, we merge this augmented dataset of unrelated personas with the raw persona dataset to eliminate the bias.

\subsection{Pre-Training Details}
We employ the augmented large-scale persona dataset as the pre-training corpora. 
To guarantee the data quality, we follow the elaborate cleaning process as PLATO-2 \cite{bao-etal-2021-plato}. 
After filtering, the data is split into training and validation sets. 
The training set contains 211M samples, and the validation set contains 0.2M samples.
We reuse the BERT-base-uncased vocabulary \cite{DevlinCLT19}. 
The maximum sequence length for the persona, dialogue context, and target response are all set to 128.
We use Adam as the optimizer with a learning rate scheduler of linear warmup and cosine decay. 
The warmup stage covers the first 30000 steps, and the peak learning rate is 5e-5.
The training of the model was done on 8 Nvidia Telsa V100 32G GPUs with a batch size of 256.

\begin{table*}
    \centering
    \resizebox{\textwidth}{!}{
    \begin{tabular}{l|ccccccc|cccc}
        \toprule 
        Method & PPL $\downarrow$ & Dist-1/2 $\uparrow$ & BS $\uparrow$ & E $\uparrow$ & N & C $\downarrow$ & CS $\uparrow$ & Flu. $\uparrow$ & Cohe. $\uparrow$ & Info. $\uparrow$ & P.C. $\uparrow$ \tabularnewline
        \midrule
        
        Baseline & 43.48 & 1.24/7.41 & 85.99 & 11.0 & 81.4 & 7.6 & 5.1 & 1.74 & 1.02 & 0.24 & -0.20 \tabularnewline
        \midrule
        
        DialoGPT & - & \textbf{5.00/21.61} & 85.23 & 8.1 & 86.3 & 5.6 & 4.5 & 1.72 & 1.16 & 0.30 & -0.12 \tabularnewline
        
        PPDS-woP & 20.19 & 3.76/19.42 & 86.00 & 11.6 & 83.0 & 5.3 & 10.7 & 1.76 & 1.26 & 0.44 & 0.02 \tabularnewline

        PPDS-woA & 18.19 & 3.19/16.67  & 86.16 & 33.8 & 58.8 & 7.4 & 41.6 & 1.74 & 1.40 & 0.88 & 0.14 \tabularnewline
        
        PPDS & 18.24 & 3.33/17.65 & 86.23 & \textbf{42.7} & 51.5 & 5.8 & \textbf{49.5} & 1.92 & 1.54 & 1.00 & 0.42 \tabularnewline
        \midrule
        
        DialoGPT-finetuned & - & 4.04/20.61 & 86.58 & 31.0 & 61.9 & 7.0 & 30.2 & 2.00 & 1.54 & 0.76 & 0.16 \tabularnewline
        
        PPDS-woP-finetuned & 15.93 & 3.10/14.86 & 86.38 & 19.2 & 75.8 & 5.0 & 18.1 & 1.98 & 1.56 & 0.56 & 0.04 \tabularnewline

        PPDS-woA-finetuned & 15.41 & 3.00/15.42 & 86.56 & 37.0 & 57.9 & 5.0 & 40.6 & 1.98 & 1.66 & 1.02 & 0.32 \tabularnewline
        
        PPDS-finetuned & \textbf{15.21} & 3.02/15.83 & \textbf{86.61} & 39.1 & 56.8 & \textbf{4.1} & 44.3 & \textbf{2.00} & \textbf{1.80} & \textbf{1.14} & \textbf{0.44} \tabularnewline
        
        \bottomrule
    \end{tabular}}
    \caption{Quantitative and human evaluation results. The best results are highlighted in bold.}
    \label{tab:Result}
\end{table*}

\section{Experiments}
\subsection{Experiment Setup}
We evaluate our models on persona dialogue generation with PERSONA-CHAT \cite{KielaWZDUS18} which is a crowd-sourced dataset covering rich persona features. The training and test sets are used for fine-tuning and evaluation, respectively.

\noindent \paragraph{Baselines.} To evaluate the performance of PPDS, the following dialogue generation models including non-fine-tuned and fine-tuned ones are compared in the experiments.

\begin{itemize}
    \item \textbf{Baseline}: Our vanilla PPDS, trained from scratch on PERSONA-CHAT without pre-training on the large-scale personal dialogue dataset.
    \item \textbf{DialoGPT}: Pre-trained on GPT-2 \cite{radford2019language} using Reddit comments. We compare its medium version, which reports the best performance \cite{ZhangSGCBGGLD20}.
    \item \textbf{DialoGPT-finetuned}: Fine-tuned DialoGPT on PERSONA-CHAT by concatenating the persona with the dialogue context.
    \item \textbf{PPDS}: Our proposed model pre-trained on the large-scale persona dialogue dataset with persona augmentation.
    \item \textbf{PPDS-woP}: Pre-trained without a persona.
    \item \textbf{PPDS-woA}: Pre-trained without persona augmentation.
    \item \textbf{PPDS-finetuned}: Our PPDS pre-trained on the large-scale persona dialogue dataset with persona augmentation and fine-tuned on the PERSONA-CHAT.
    \item \textbf{PPDS-woP-finetuned}: PPDS fine-tuned on PERSONA-CHAT.
    \item \textbf{PPDS-woA-finetuned}: Pre-trained without persona augmentation and fine-tuned on PERSONA-CHAT.
\end{itemize}

\subsection{Evaluation Metrics}
We evaluate the response quality and persona consistency of the personal dialogue generation through both quantitative and human evaluations.
For dialogue quality, we follow common practice \cite{KielaWZDUS18} to employ the following quantitative metrics: 
(1) Perplexity \textbf{(PPL)}. Lower perplexity means better language modeling.
(2) Distinct 1/2 \textbf{(Dist-1/2)} \cite{LiGBGD16} denotes the ratio of distinct uni-grams/bi-grams, where higher distinct means better response diversity. 
(3) BertScore \textbf{(BS)} \cite{ZhangKWWA20} measures the coherence similarity between predicted response and target response measured through the BERT model.
For persona consistency, we employ the ratios of responses that are entailed (\textbf{E}), neutral (\textbf{N}), and contradicted (\textbf{C}) to the personas, which are measured by an NLI model.
We also calculate Consistency Score (\textbf{CS}) \cite{MadottoLWF19} to measure persona consistency, which summarizes the result of NLI as follows:

\begin{align}
    NLI(\boldsymbol{R},\boldsymbol{P}_{i}) &=
    \begin{cases}
        -1, & if\ \boldsymbol{R}\ contradicts\ \boldsymbol{P}_{i},\\
        0, & if\ \boldsymbol{R}\ is\ neutral\ to\ \boldsymbol{P}_{i},\\
        1, & if\ \boldsymbol{R}\ entails\ \boldsymbol{P}_{i}.
    \end{cases} \\
    CS(\boldsymbol{R}) &= \sum_{i=1}^{M}NLI(\boldsymbol{R},\boldsymbol{P}_{i})
\end{align}

The NLI model is fine-tuned on the DNLI dataset \cite{WelleckWSC19} based on the pre-trained large RoBERTa model \cite{LiuOGDJCLLZA19}, achieving test set accuracy of $93.3\%$ on DNLI.

As suggested in the empirical study \cite{LiuLSNCP16}, the correlation between quantitative metrics and human judgments may be weak in open-domain dialogue generation. 
Therefore, we also conduct human evaluations in the experiments. 
Crowd-sourcing workers who are proficient in dialogue tasks are asked to evaluate the responses on the following metrics: (1) Fluency (\textbf{Flu.}) measuring whether the response is fluent and grammatically correct.
(2) Coherence (\textbf{Cohe.}) measuring whether the response is relevant and consistent with the context.
(3) Informativeness (\textbf{Info.}) evaluating whether the response is informative or not given the context.
(4) Persona Consistency (\textbf{P.C.}) checking whether the response has conflicts with the persona, where $1$ means persona-related and consistent, $0$ means neutral, and $-1$ means contradicted.
Fluency, Coherence, and Informativeness are all rated on a scale of [0, 1, 2], with higher scores indicating better quality.

\begin{table}[htbp]
    \center
    \small
    \resizebox{\columnwidth}{!}{
    \begin{tabular}{l|l}
        \toprule 
        Persona & I like to drink wine \tabularnewline
        
        \midrule
        
        Context & ... \tabularnewline
        
        & A: Which do you prefer white or red? \tabularnewline
        
        & B: Fermented. I prefer well-aged and fermented. \tabularnewline
        
        & \ \ \ \ \ I could bathe in it! \tabularnewline
        
        & A: Good choice. I always like a nice dry white wine. \tabularnewline
        
        
        \midrule
        
        PPDS-woP & My secret is that \textcolor{red!60!gray}{I don't drink}. I don't know why. \tabularnewline
        
        \midrule
        
        \textbf{PPDS} & Me too. My body would be so strong if I had a \textcolor{green!40!gray}{dry wine}! \tabularnewline
        
        \bottomrule
    \end{tabular}}
    \caption{A qualitative example of a persona-consistent response generated by our model (PPDS) compared to a model pre-trained without a persona (PPDS-woP).}
    \label{tab:Case}
\end{table}

\subsection{Evaluation Result}

The evaluation results are summarized in Table \ref{tab:Result}. The baseline model struggles to perform well due to the limited scale of the PERSONA-CHAT dataset, which is inadequate for its large parameters. In contrast, all methods with large-scale pre-training generate fluent and coherent responses, demonstrating the benefits of such pre-training. However, DialoGPT and PPDS-woP exhibit poor persona consistency with low $CS$ and $P.S.$ scores. Although fine-tuning improves the persona consistency of DialoGPT-finetuned and PPDS-woP-finetuned, the performance remains unsatisfactory.

Our PPDS-woA and PPDS models, pre-trained on the large-scale persona dialogue dataset, achieve significantly better persona consistency scores, surpassing both PPDS-woP-finetuned and DialoGPT-finetuned by a large margin. This indicates that large-scale pre-training on persona dialogue data can greatly enhance the persona consistency of dialogue models. Additionally, with our proposed persona augmentation, PPDS demonstrates superior persona consistency and response quality in both quantitative and human evaluations compared to PPDS-woA, confirming the effectiveness of persona augmentation in mitigating bias in the constructed dataset. The improvements in reducing contradictions and enhancing coherence and informativeness are particularly notable. Ultimately, through pre-training on the large-scale persona dialogue dataset with persona augmentation and subsequent fine-tuning on the PERSONA-CHAT dataset, our PPDS-finetuned achieves the highest scores in most quantitative and human evaluations, showcasing its superior understanding of persona consistency.  A qualitative example is presented in Table \ref{tab:Case} to further illustrate the effectiveness of our model in maintaining persona consistency.


\section{Conclusion}
\label{sec:con}

In this work, we introduce a summarization-based persona extraction model to construct a large-scale persona dialogue dataset. Based on the dataset, we propose PPDS, an open-domain persona dialogue system that leverages large-scale pre-training for achieving persona consistency in dialogue generation. Both quantitative and qualitative evaluations demonstrate the effectiveness of our approach. Given that the experiments were conducted with relatively cost-efficient models (T5 and BERT) and still yielded promising results, this work encourages future research to apply these techniques in building large-scale dialogue models and enhancing dialogue generation systems for industrial applications.

Beyond the discussed techniques, we also encourage exploration towards the following directions in constructing better persona datasets and training persona-consistent dialogue models for different application scenarios. First, the recent emergence of LLM-in-the-loop methodologies \cite{hongposition} offers a promising path by incorporating the natural language understanding capabilities of LLMs to enhance the persona extraction process for complex, compositional personas. Second, the source of extraction can extend from textual data to multimodal datasets, particularly conversational speech that contains rich persona information \cite{10.1145/3394171.3414392, 10.1145/3514221.3520158}. Lastly, extending the persona from individual behaviors to larger entities, such as brand personality \cite{aaker1997dimensions}, would further enhance the practical value of the proposed methods in various downstream domains, such as the hospitality and service sectors \cite{ng2024function}.


\bibliography{Reference}


\end{document}